# Prompt Selection and Augmentation for Few Examples Code Generation in Large Language Model and its Application in Robotics Control

On Tai Wu, Frodo Kin Sun Chan, Zunhao Zhang, Yan Nei Law, Benny Drescher and Edmond Shiao Bun Lai*

## Abstract

Few-shot prompting and step-by-step reasoning have enhanced the capabilities of Large Language Models (LLMs) in tackling complex tasks including code generation. In this paper, we introduce a prompt selection and augmentation algorithm aimed at improving mathematical reasoning and robot arm operations. Our approach incorporates a multi-stage example augmentation scheme combined with an example selection scheme. This algorithm improves LLM performance by selecting a set of examples that increase diversity, minimize redundancy, and increase relevance to the question. When combined with the Program-of-Thought prompting, our algorithm demonstrates an improvement in performance on the GSM8K and SVAMP benchmarks, with increases of 0.3% and 1.1% respectively. Furthermore, in simulated tabletop environments, our algorithm surpasses the Code-as-Policies approach by achieving a 3.4% increase in successful task completions and a decrease of over 70% in the number of examples used. Its ability to discard examples that contribute little to solving the problem reduces the inferencing time of an LLM-powered robotics system. This algorithm also offers important benefits for industrial process automation by streamlining the development and deployment process, reducing manual programming effort, and enhancing code reusability.

## 1. Introduction

Industrial process automation plays a crucial role in supporting diverse production processes. Real-time embedded automation controllers are at the core of these processes, efficiently managing large volumes of sensor data. They execute control logic based on this data and generate output signals to actuate machines. Robot arm operations, in particular, are widely employed in manufacturing, logistics, and warehouse management. Automating the code generation for robot arm control offers numerous benefits, including enhanced efficiency and cost savings. In a robot arm sorting application, real-time embedded automation controllers are responsible for processing sensor data, executing control algorithms, and generating precise movements for the robot arm. By automating the code generation process, engineers can streamline development cycles and reduce the manual effort required to program the controllers. Automated code generation simplifies the development process by allowing engineers to specify high-level requirements and have the corresponding control code automatically generated. This reduces the need for manual coding, minimizes human errors, and accelerates the overall development timeline.

Since code generation holds significant advantages, it leads researchers to explore various approaches in the past. One promising approach that has garnered attention is the use of Large Language Models (LLMs). However, despite their potential, there is still ample room for improvement in this area. Prompt engineering has been pivotal in improving the performance of LLMs for complex problems. Normally, the prompts include but not limited to the instructions, which show the tasks or guidelines for LLM to perform, and examples, which demonstrate how to response with different user queries. The development of prompting strategies that lead LLMs through multi-step reasoning processes is a significant advancement in this field. These methods have been innovative; in particular, the Chain-of-Thought (CoT) prompting (Wei *et al.*, 2022) has been shown to improve LLM performance dramatically with a small number of examples. Even with these developments, there is still a need for more effective techniques for the selection and enhancement of prompts, which can help LLMs' reasoning skills even further. The challenge lies in obtaining general selection criteria from mostly empirical observations (Fu *et al.*, 2022).

To address this, we present a new systematic algorithm that directly computes and assesses a wide range of metrics extracted from in-context examples, hence improving the prompt selection procedure. Our algorithm is characterized by three main stages: expanding the existing in-context examples by example augmentation, measuring the usefulness of each example using a learned scoring system, and eliminating examples with low score to reduce the number of prompts (refer to Figure 1). This approach is intended to be computationally and data-efficient, requiring fewer examples.

We evaluate the performance of our algorithm rigorously using the Program-of-Thought (PoT) prompting approach (W. Chen *et al.*, 2023a) on multiple mathematical reasoning exams including SVAMP (Patel, Bhattamishra and Goyal, 2021a) and GSM8K (Cobbe *et al.*, 2021a). We further investigate its use in robotics, putting our algorithm into practice in a pick and place system with the Code as Policies (CaP) architecture (Liang *et al.*, 2022) and assessing its usefulness in robotic control tasks.

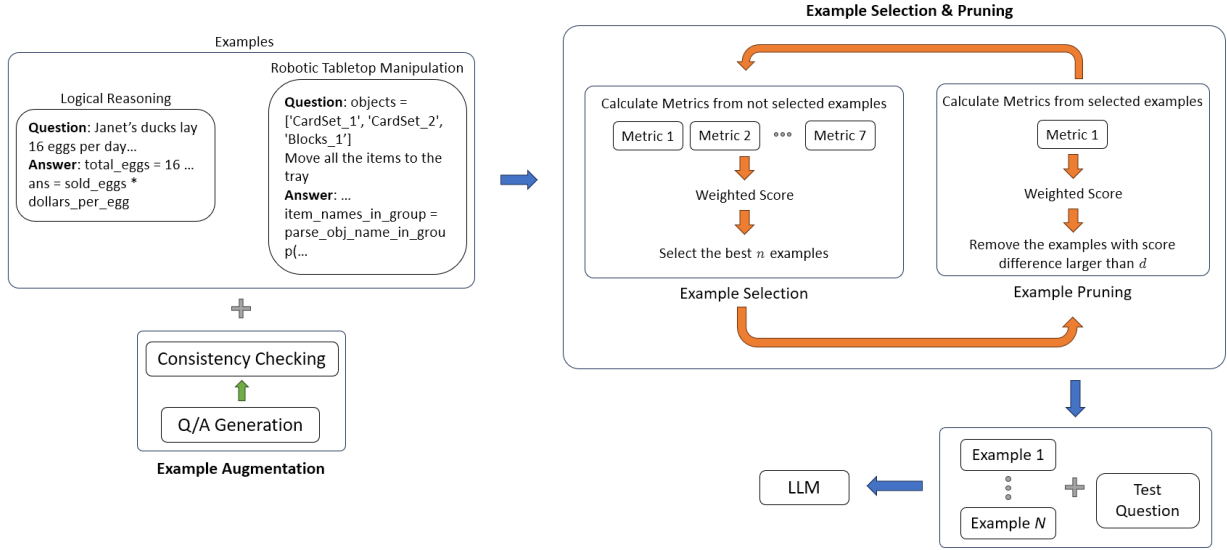

Figure 1: A brief overview of our approach, each section is illustrated with a different color. **Example Augmentation**: **(1)** Q/A Generation, question and answer are rephrased by an LLM, of which the modified part is highlighted in yellow **(2)** Consistency Checking, generated answer is verified against the original answer before using them (see Section 3.1). **Example Selection & Pruning**: **(1)** Compute a weighted score base on the selection metrics and select the best scored examples. **(2)** Compute a weighted score base on the pruning metrics, and prune the worst scored example if the difference to the second worst is larger than *d*. Finally, the *N* selected examples are combined with the test question to prompt the LLM. (see Section 3.2)

## 2. Related Work

### 2.1 Robotics Programming with LLMs

Recent advancements have explored the use of LLM for robotic task programming, Singh *et al.*'s (2022) PROGPROMPT and Zeng *et al.*'s (2022a) Socratic Models demonstrate LLMs' capabilities in generating task plans and multimodal reasoning. Li *et al.*'s (2023) Chain of Code and Hao *et al.*'s (2023) Reasoning via Planning further extend LLM applications into semantic task reasoning and strategic planning exploration. However, these studies do not delve into optimizing example selection for LLMs in robotics, leaving a research gap our paper addresses by proposing a method that enhances LLM efficiency in robotic task execution through targeted example selection.

### 2.2 Multi-Step Reasoning

A major turning point was reached with LLM prompting with the development of Chain-of-Thought (CoT) prompting (Wei *et al.*, 2022), which showed that few-shot prompting could significantly improve LLM performance. Through step-by-step guidance in information processing, this method effectively enhances LLMs' capacity to handle increasingly sophisticated reasoning tasks. Adding to this, there are several methods that uses the code interpreter to guide the LLMs' reasoning steps (Gao *et al.*, 2023a; W. Chen *et al.*, 2023a). The Program-of-Thought (PoT) method (W. Chen *et al.*, 2023a) in particular, has shown to outperform CoT in mathematical reasoning tasks significantly. It separates the reasoning process from computational tasks by implementing code interpreters for execution. By allowing LLMs to concentrate on logical progression, this separation greatly reduces errors and establishes a new standard in prompt engineering for complex problem-solving. We are utilizing PoT due to its better performance in code synthesis. Additionally, we would like to leverage the code nature of the in-context examples to compute the code semantic similarities between example questions and test questions (Ren *et al.*, 2020; Yu *et al.*, 2022b).

## 2.3 Example Selection for Prompting

Examples are painstakingly handcrafted by humans, as it requires a careful selection of questions and high-quality reasoning steps. Studies has also observed that the quality of examples can have significant effects on the quality of response from LLMs, in particular, order sensitivity, complexity, diversity, style sensitivity, are very important to performance (KaShun Shum, Diao and Zhang, 2023). As a result, automatic prompt searching is developed to improve LLM performance (Rubin, Herzig and Berant, 2022a; Zhang *et al.*, 2022a; KaShun Shum, Diao and Zhang, 2023; Pitis *et al.*, 2023a). However, these algorithms often use a limited number of features or proxy to measure the effectiveness of each prompt. Hence, our work hence aims to use all of the known effective features adjusted according to the weights ascertained via Bayesian optimization (Bergstra *et al.*, 2011a; Lacoste *et al.*, 2014; Watanabe, 2023) in our search algorithm.

## 3. Proposed Algorithm

Our goal is to construct prompt $P$, that when given to an LLM, would provide maximum likelihood to generate the correct/desire outcome. Given in-context examples with questions $Q = \{q_1, q_2, \ldots, q_i\}$ and reasoning chains $A = \{a_1, a_2, \ldots, a_i\}$, we first expand $q_i, a_i$ with a three-stage augmentation process using an LLM. The answer of the solution is obtained by executing the program reasoning chains and performed Consistency Checking on the generated $q'_i, a'_i$ against the original $a_i$ to check if it gives the same solution. The prompt augmentation is further discussed in Section 3.1. The $q'_i, a'_i$ that has passed Consistency Checking is added into the expanded questions and answer set, $Q', A'$. We can then construct $P$ with a subset of $Q', A'$ selected based on specific criteria, which is further discussed in Section 3.2.

Our selection scheme is an iterative process that navigates through the example space, and dynamically adjusting the composition of the prompt set. This approach is similar to the classic feature selection algorithm Maximum Relevance — Minimum Redundancy (Ding and Peng, 2003). This function leverages embeddings of questions and answers, alongside a suite of metrics that assess complexity, semantic similarity, and concept overlap to score and select the most effective examples. The scheme refines the pool of potential prompts through a cycle of evaluation and selection, ensuring each iteration moves closer to the ideal set of examples that will guide the LLM towards generating optimal responses.

### 3.1. Example Augmentation

Using examples when prompting an LLM aims to guide it effectively without requiring fine-tuning. Initially, users supply basic (seed) examples, but it's often time consuming and challenging to craft a lot of diverse and complex seed examples. To address these limitations, our scheme generates new examples via an LLM in a step by step augmentation process, incorporating consistency checking to verify the accuracy of these examples.

*Consistency checking*

Consistency checking involves comparing the generated answer with the original to assess accuracy. This comparison occurs in two stages: first, evaluating the correctness of the answer by executing and comparing the outputs of the answer steps. If discrepancies arise, they are deemed inconsistent. Second, if the output match, the similarity between the steps is compared since different intermediate steps can lead to the same result. This comparison utilizes the embeddings from the answer steps, using the cosine similarity and the embedding service provided by the Gemini API.

*Question and Answer Generation*

Our augmentation process generates new question and answer pairs from seed examples. To ensure the newly generated answers are accurate and consistent, Shao *et al*. (2023a) proposed the method by prompting an LLM with a reversed question-answer sequence. The LLM would then generate the answer first, then the corresponding question. After generating the new answer, we verify the question-answer pair by comparing the newly generated answer and the original answer. Apart from the question generation, the augmentation is also improved by increasing the complexity of the answer (Fu *et al.*, 2022), which is defined by the answer sentence length. By combining the question generation, the question-and-answer modification is also proposed in this paper for increasing the complexity of the question and answer.

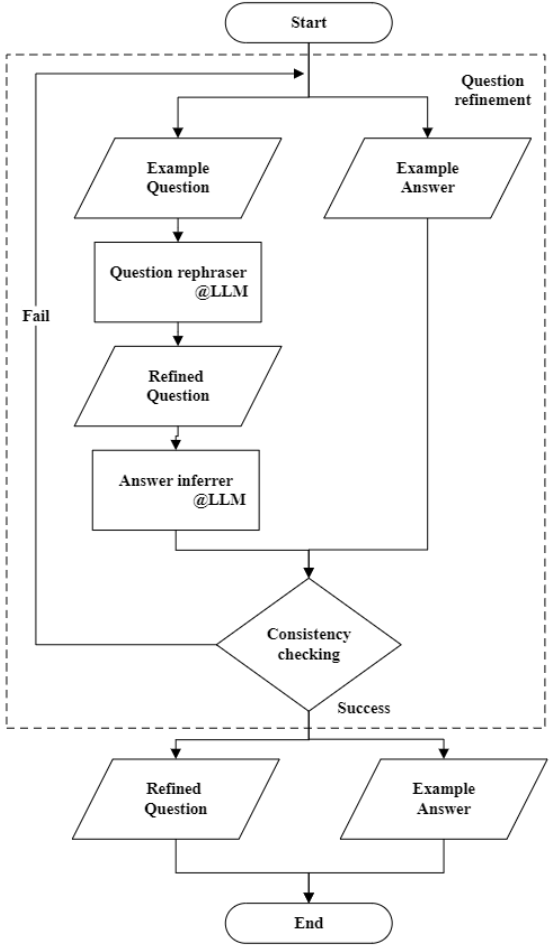 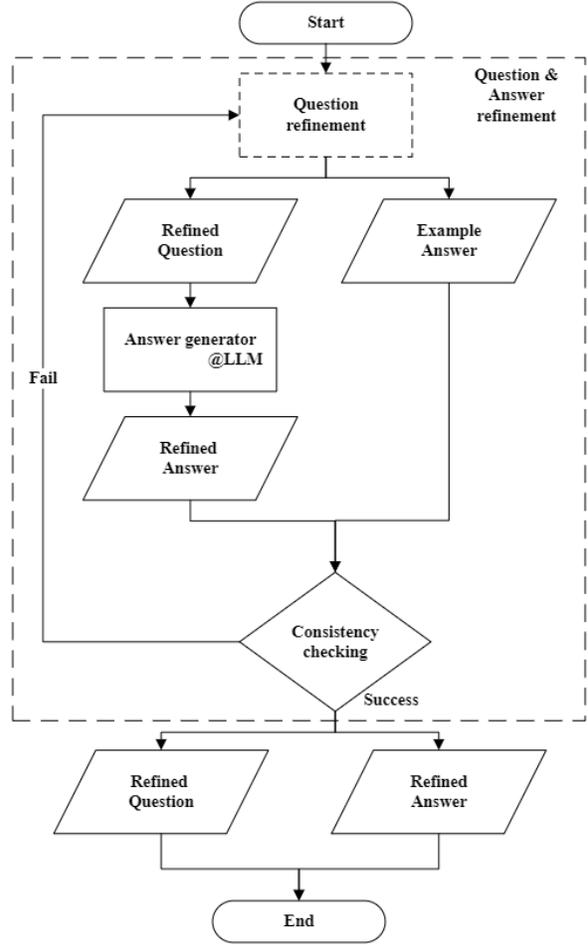

(a)                  (b)

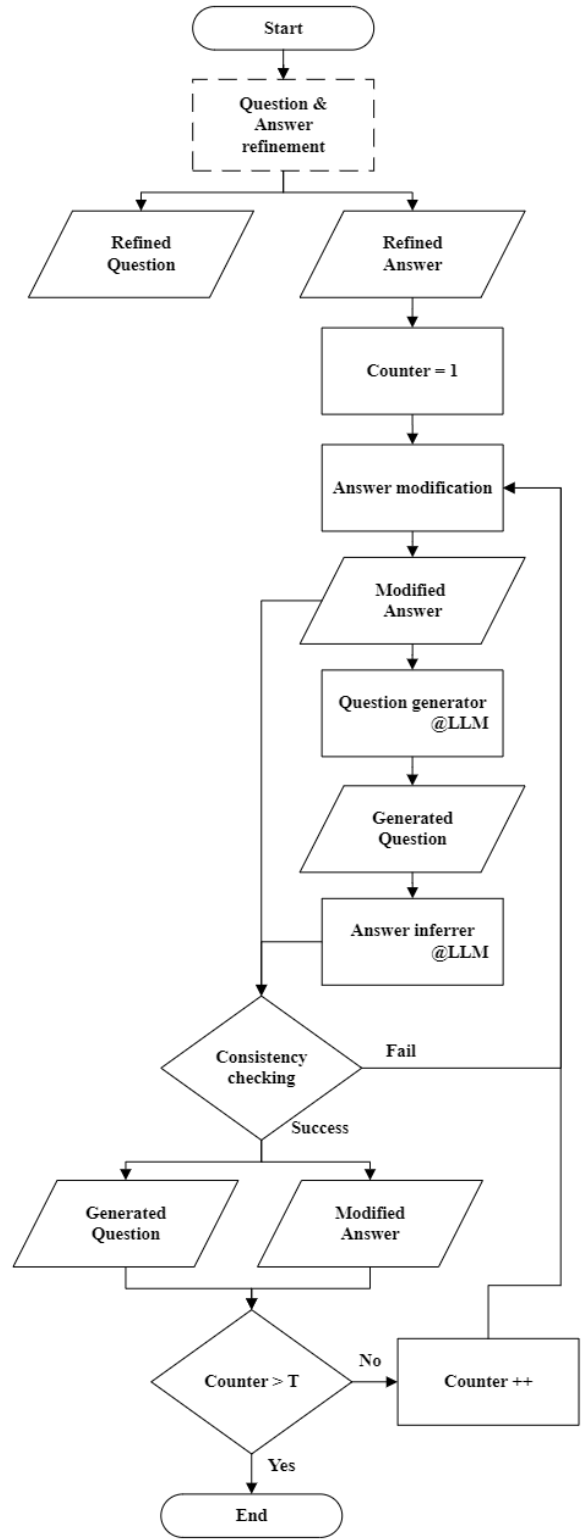

(c)

Figure 2: The control flow diagram of the three-stage example augmentation. (a) The first stage augmentation, (b) the second stage augmentation and (c) the third stage augmentation with T-iteration. T is the number of iterations for modifying the answer.

The example augmentation process is divided into three stages. The first stage involves rephrasing the question of an example using the "Question Rephraser" module, which operates through an LLM with a designated prompt, as illustrated in Fig. 3(a). As shown in the example, the question-answer pair is first presented, followed by an instruction on rephrasing the question. Fig. 2(a) shows control flow diagram of the first stage augmentation. After the refined question is formed, it is used to generate an answer with intermediate steps using the "Answer Inferrer", as shown in Fig. 3(b). Both the original question-answer pair, and the newly rephrased question is presented to an LLM, and a new answer is refined by an LLM. Finally, we perform consistency checking on the newly inferred. If it passes, the rephrased question and answer are accepted and kept. If not, the process is restarted until we generate an answer that's consistent.

For the second stage, the refined question and example answer are first obtained from the first stage. Fig. 2(b) shows its control flow diagram. By using the refined question, an answer is generated by the Answer generator and the prompt example is shown Fig. 3(c). By using the Answer Generator, the generated answer is not referencing the original seed example, but rather a different seed example. It is therefore not limited to the original seed example's answer to increase its diversity. Then, the consistency checking is applied to the generated answer and the refined answer. If it passes, the generated answer and the refined question are kept for use. If not, the process can restart again or abort.

For the third stage, it is designed to modify the question and answer instead of refining the question and answer in the first two stages, while keeping it from being too dissimilar from the seed examples. Thus, consistency checking is still required to ensure the correctness of the question and answer. Fig. 2(c) shows its control flow diagram. Firstly, the refined question and answer are obtained from the second stage. Then, the answer modification is applied to change the refined answer. For a numerical answer, the answer modification is performed by using basic operations including addition, subtraction, multiplication, and division. After the forming the modified answer, the newly generated question and answer are extracted by the Question Generator, which creates a question from the modified answer and a prompt used is shown in Fig. 3(d). The Answer Inferrer used is similar to that of the first stage, except that the prompt is formed by the refined question and answer. If the generated question and modified answer passes the consistency checking, they are good for use. If not, the process can restart again or abort. To further increase the complexity of the question and answer, the generated question and the modified answer can be reused for the third stage augmentation.

Figure 3: The prompt examples of question-and-answer generators. (a) Question Rephraser: the prompt to refine question, (b) Answer Inferrer: the prompt to refine answer, (c) Answer Generator: the prompt to modify answer, (d) Question Generator: the prompt to modify question. A set of examples for inference reference is highlighted in blue and the modified steps in the answers are highlighted in yellow.

### 3.2. Examples Selection & Pruning

To effectively select and prune the examples for prompt construction, the following factors (metrics) are proposed for measurement of different examples on different tasks.

- **Complexity**: We adopt the metric of complexity as suggested by Fu et al. (2022), where complex prompts are associated with improved LLM performance. Complexity is quantified by indicators such as the number of line breaks "\n" within an example, serving as a proxy for the number of reasoning steps or the depth of thought required.
- **Semantic Similarity** (Relevance/Similarity): This metric assesses how closely the content of an example aligns semantically with the test question, ensuring the selected prompts are contextually relevant to the task. Previous methods have used semantic similarities as well to select examples (Rubin, Herzig and Berant, 2022a), but we have expanded our similarity to the code semantic similarity.

- **Concept Similarity**: To capture nuances beyond semantic similarity, we evaluate the conceptual overlap between questions using an LLM. This metric is particularly useful when two questions may be semantically similar but differ in their underlying ideas or problem structures.

---

**Scheme 1** Examples Selection

---

**Parameters:** Weights for metrics $w$, Convergence threshold $\epsilon$, Acceptable difference $\delta$
**Input:** Example Questions $Q$, Example Answers $A$, Test Question $q_t$
**Output:** Chosen Example $C$

    Initialize $C \leftarrow \emptyset$, Example Candidates $C' \leftarrow \{(q_{j_1}, a_{j_1}), \ldots, (q_{j_N}, a_{j_N})\}$ of length $N$
    **for** each iteration **do**
        **for** each $(q, a)$ in $C'$ **do**
            $m_{q,a,q_t} \leftarrow$ CalculateMetrics$(q, a, Q_t)$
            $S$.insert$(w \cdot m_{q,a,q_t})$                                   # $m$ and $w$ are a $1 \times 5$ vector
        **end for**
        $s \leftarrow argmax(S)$                                                     # Find the best-scoring example
        $C$.insert$(s)$
        $C'$.remove$(s)$
        **if** |previous max score $-$ max $(s)| < \epsilon$ **then**
            **break**
        **end if**
        $R \leftarrow$ Similarity between Chosen Example $C$
        $R$.sortDescending()
        **if** $R_0 - R_1 > \delta$ **then**
            $C$.remove$(argmax(R))$
        **end if**
    **end for**

---

| |
|---|
| **Function** CalculateMetrics() |
| **Input:** Example Question $Q$, Example Answer $A$, Test Question $Q_t$ |
| **Output:** Metrics $m$ |
| relevance $\leftarrow S_C(Encode(Q), Encode(Q_t))$     # $S_C$ is the cosine similarity, and encoded using Sentence BERT (Reimers and Gurevych) |
| concept $\leftarrow S_C(LLM(A), LLM(Q))$     # Using LLM to generate concepts for $Q, A$ |
| $A_c \leftarrow Complexity(A)$     # Number of line breaks in $A$ |
| complexity $\leftarrow norm(A_c)$ |
| similarity $\leftarrow S_C(Encode(A), Encode(LLM(Q_t)))$     # Using zero-shot prompting to generate rough answer |
| $m \leftarrow$ [relevance, concept, complexity, similarity] |

The scheme begins by initializing two sets: chosen and non-chosen examples. As the scheme iterates, each not-yet-selected example is evaluated against the test question using the metrics described above. Examples are scored based on a weighted sum of these metrics, with the weights determined through Bayesian optimization to fine-tune these weights based on the performance on a subset of the GSM8K training dataset.

At each iteration, the example with the highest score is moved from the non-chosen set to the chosen set. To ensure the diversity of chosen exemplars, the scheme compares the average similarities of each example answer to others, and pruning the example that are most similar to the others based on a predefined acceptability threshold for redundancy. By focusing on removing examples that are significantly more redundant, the process refines the set to maintain a wide-ranging and informative collection of examples.

The iterative process continues until a convergence criterion is met, which is either until the desired set size is met, or until an improvement of the weighted score that is smaller than a threshold.

By maintaining balance between diversity and similarity, the selected examples enable the LLM to have a well-rounded understanding of the task, thereby maximizing its problem-solving ability.

## 4. Experimental Results

To evaluate the effectiveness of our proposed algorithm, we chose the two math dataset GSM8K and SVAMP, which is public available for evaluation. From the PoT repository, we can reuse the prompts used in the original evaluation and from those apply our prompt selection and augmentation algorithm.

To finetune our weights *w* for different metrics, we used a subset of the training set (200 Q,A pairs) of GSM8K for tuning. We have chosen Bayesian Optimization (Bergstra *et al.*, 2011a) due to the expensive inferencing nature of LLM, with the additional evidence provide by Chen *et al.* (2023) that Bayesian Optimization is suitable for prompt selection.

## 4.1. Overall Results

Table 1: Accuracy on GSM8K and SVAMP

| Model | | GSM8K | No. Examples | SVAMP | No. Examples |
|---|---|---|---|---|---|
| Gemini Pro | PoT | 76.6% | 9 | 85.7% | 7 |
| | **PoT + Proposed Algorithm** | **76.9%** (+0.3%) | **3.98** (-55.8%) | **86.7%** (+1.0%) | **3.26** (-53.4%) |
| GPT-3.5-turbo-instruct | PoT | 73.0% | 9 | **77.8%** | 7 |
| | **PoT + Proposed Algorithm** | **74.1%** (+1.1%) | **3.98** (-55.8%) | 77.5% | **3.26** (-53.4%) |

In this study, we evaluated the performance of our proposed algorithm using the GSM8K and SVAMP datasets, which consist of 1310 and 1000 question-answer (QA) pairs respectively (excluding the examples in the prompts), based on math word problems. We used the PoT (W. Chen *et al.*, 2023a) few-shot prompting examples as seed examples for augmentation and selection. Our algorithm, applied to both Gemini Pro and GPT-3.5-turbo-instruct models, showed improvements in accuracy for both datasets, with the exception for GPT-3.5 on SVAMP. Notably, on average, we selected 3.98 and 3.26 examples from the augmented sets for GSM8K and SVAMP respectively, as compared to the original 9 and 7 examples from the PoT testing shown in Table 1, demonstrating the efficiency and impact of our augmentation and selection algorithm in enhancing LLM performance in mathematical reasoning tasks.

## 4.2. Augmentation Evaluation

To further explore the effectiveness of the augmented examples, we used the SVAMP dataset for the evaluation. Each example generated from PoT underwent the three-stage augmentation process. Subsequently, augmented examples were paired with their original counterparts for LLM inference in a two-shot manner, with the average accuracy of these pairings presented in Table 2. For the first two stages of augmentation, the question and answer were refined to create new example. In subsequent stages of augmentation, the LLM generates more complex examples with each new iteration (increasing the question and answer length), as detailed in Fig. 2(c). The LLM used for this evaluation was Gemini 1.0 Pro.

In Table 2, each column represents a different seed example, and each row represents the stage of augmentation. The rows with the third stage augmentation have different maximum number of iterations for independent testing. After obtaining the answer steps by inferencing with the LLM, the answer is then calculated and compared with the ground truth. Thus, the average accuracy is calculated as the amount of correctness over the number of QA pairs for each dataset. On the other hand, the example generation in augmentation is not always successful because LLM may give the wrong inference. Thus, the augmented examples are not available because the augmentation fails on consistency checking.

In Table 2, the original example with either the first or second stage augmentation performs better than the baseline in all the examples. In addition, the original example with the third stage augmentation also has better accuracy than the baseline for all the examples and the largest difference is more than 5% in Example S2 and S3. From the experimental results, they show that the proposed augmentation scheme not only provides more examples, but also enhances the LLM inference with different examples.

Table 2 The experimental result on SVAMP dataset with proposed data augmentation scheme

|  | Example S1 | Example S2 | Example S3 | Example S4 | Example S5 | Example S6 | Example S7 |
|---|---|---|---|---|---|---|---|
| **Given example + Given example (baseline)** | 80.90% | 78.00% | 78.77% | 79.95% | 80.64% | 80.40% | 80.40% |
| **Given example + Stage 1 augmentation** | 83.00% | 83.30% | 83.17% | 83.10% | 82.82% | 82.97% | 83.06% |
| **Given example + Stage 2 augmentation** | 82.80% | 82.40% | 83.37% | 83.53% | 83.40% | 83.07% | **83.09%** |
| **Given example + Stage 3 augmentation with 1 iteration** | 81.10% | 82.10% | 82.80% | N/A* | 82.97% | 83.06% | N/A* |
| **Given example + Stage 3 augmentation with 2 iterations** | 83.10% | 82.95% | N/A* | 83.07% | 82.55% | 79.58% | N/A* |
| **Given example + Stage 3 augmentation with 3 iterations** | 82.90% | N/A* | N/A* | 83.45% | 83.47% | 83.40% | N/A* |
| **Given example + Stage 3 augmentation with 4 iterations** | 82.00% | 83.00% | 82.93% | N/A* | 82.75% | 82.82% | N/A* |
| **Given example + Stage 3 augmentation with 5 iterations** | **83.80%** | **83.35%** | **84.30%** | **84.68%** | **84.12%** | **83.98%** | N/A* |

*The augmented example is not available because the augmentation fails on consistency checking

### 4.3. Tabletop Manipulation Simulation Evaluations

We evaluate our prompt selection algorithm in a simulated tabletop environment, which is shown in Fig. 4(a), similar to the environment used in CaP (Liang *et al.*, 2022). The setup used a simulated UR5E robot working in an environment with different colored bowls and blocks. Note that Gemini Pro was used as LLM in this evaluation. The LLM is first given 13 initial tabletop manipulation prompts, using a variety of attributes. Then the LLM is given 6 unseen instructions (UI) with unseen attributes (UA), with each instruction having 20 variety of different attributes and different initial tabletop configurations. For each trial out of $6 \times 20 = 120$ combinations, we used our proposed algorithm to select the best prompts out of the 39 prompts (13 original + 13 the first stage augmentation + 13 the second stage augmentation), and we observed an increase in successful completed task compared to using all 13 original examples. The average number of examples used is also decreased to 3.675. The corresponding results can be found in Table 3.

We have also tested our algorithm in a real world tabletop manipulation, as shown in Fig. 4(b). The details of the setup and demo can be found on our website[2].

Table 3: Accuracy on Unseen Attributes and Instructions in Tabletop Simulation

|  | PyBullet Tabletop UA/UI | Average no. of examples used |
|---|---|---|
| CaP | 45.8% | 13.0 |
| **CaP + Proposed Algorithm** | **49.2% (+3.4%)** | **3.675 (-71.7%)** |

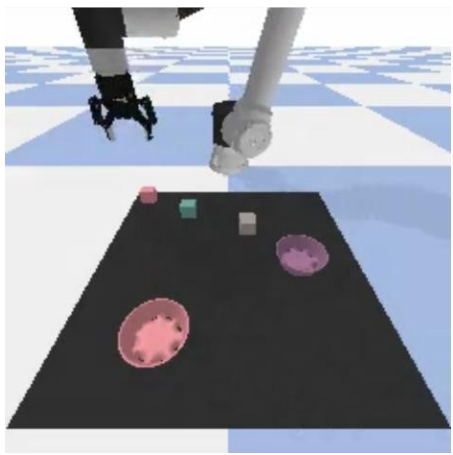
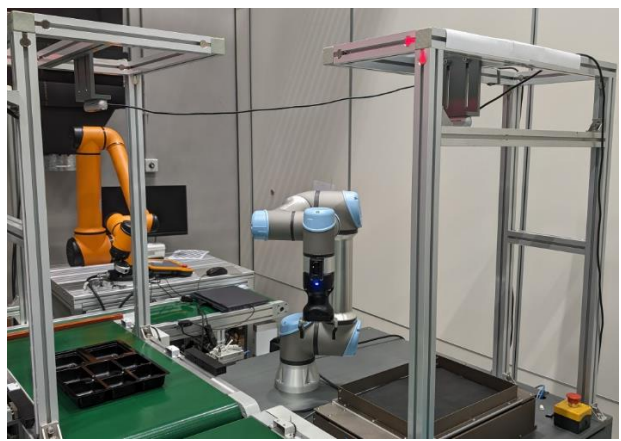

**(a)**            **(b)**

Figure 4: **(a)** simulation **(b)** real world environment in Section 4.3 and 4.4

## 5. Conclusion

In this work, we presented an innovative algorithm designed to optimize prompt selection and augmentation for LLM, with a focus on improving robotics control tasks. Our approach, which combines a multi-stage example augmentation process with a strategic selection mechanism based on a comprehensive set of metrics, demonstrates enhancements in LLM performance for both mathematical reasoning and robotic control applications. By verifying with Gemini Pro and GPT-3.5-turbo-instruct models, two public datasets (GSM8K, SVAMP), a public simulation setup (CaP tabletop tasks) and a real environment tabletop setup (self-defined tasks) in the experiments, it shows that the proposed algorithm not only improves the accuracy of the LLM performance in different applications, but it also reduces the number of examples in the construction of the prompts. Thus, the proposed algorithm is beneficial for industrial process automation with LLM. Automating code generation streamlines development cycles and reduces manual programming effort for controllers. It accelerates development, minimizes errors, and promotes code reusability.

---

[2] https://hkflair-f0086.github.io/prompt-selection-augmentation/

## Acknowledgement

This work was supported by the AIR@InnoHK research cluster of Innovation and Technology Commission (ITC) of the HKSAR Government. The results presented in this paper are within the scope of the research project "Recognition and Positioning Robotic System with Vision for Sorting Application" and have come to a successful completion thanks to the support from ITC. The authors would like to express their sincere gratitude to them.